\documentclass[10pt, a4paper]{article}

\usepackage[]{lrec-coling2024} 
\usepackage{xurl}
\usepackage{times}
\usepackage{latexsym}
\usepackage[T1]{fontenc}
\usepackage[utf8]{inputenc}
\usepackage{microtype}
\usepackage{amsmath}
\usepackage{multirow}
\usepackage{array}
\usepackage{booktabs}
\usepackage{tabularx}
\usepackage{graphicx}
\usepackage[para,online,flushleft]{threeparttable}
\usepackage{tikz}
\usepackage{listings}
\usepackage{soul}
\usepackage{framed}
\usepackage{enumitem}
\usepackage{algorithm}
\usetikzlibrary{arrows,shapes.geometric,arrows.meta,bending,chains,backgrounds,fit}

\lstnewenvironment{verbatim}{\lstset{breaklines,basicstyle=\ttfamily}}{}

\newcolumntype{Y}{>{\centering\arraybackslash}X}

\title{Stance Reasoner:\\Zero-Shot Stance Detection on Social Media\\with Explicit Reasoning}

\name{Maksym Taranukhin$^1$~~~Vered Shwartz$^{2,3}$~~~Evangelos Milios$^1$}

\address{$^1$ Faculty of Computer Science, Dalhousie University\\
$^2$ Department of Computer Science, University of British Columbia\\
$^3$ Vector Institute for AI\\
{\tt \{m.t,eem\}@dal.ca, vshwartz@cs.ubc.ca}}

\abstract{
Social media platforms are rich sources of opinionated content. Stance detection allows the automatic extraction of users' opinions on various topics from such content. We focus on zero-shot stance detection, where the model's success relies on (a) having knowledge about the target topic; and (b) learning general reasoning strategies that can be employed for new topics. We present Stance Reasoner, an approach to zero-shot stance detection on social media that leverages explicit reasoning over background knowledge to guide the model's inference about the document's stance on a target. Specifically, our method uses a pre-trained language model as a source of world knowledge, with the chain-of-thought in-context learning approach to generate intermediate reasoning steps. Stance Reasoner outperforms the current state-of-the-art models on 3 Twitter datasets, including fully supervised models. It can better generalize across targets, while at the same time providing explicit and interpretable explanations for its predictions.
\\ \newline 
}

\begin{document}

\maketitleabstract

\section{Introduction}
\label{sec:introduction}

With an abundance of opinions expressed on the Internet every day, \emph{stance detection}, which aims at identifying the stance of a text towards a target of interest (an entity, claim, topic, etc.), has attracted much attention from the NLP community, as a test-bed for automatically extracting opinionated information from massive amounts of text \cite{alturayeifSystematicReviewMachine2023}. In this paper, we focus on the challenging variant of the task, \emph{zero-shot stance detection},\footnote{We use the term ``zero-shot'' to describe the evaluation of the model on targets not seen during training. This is distinct from the conventional use of ``zero-shot'' to denote unsupervised methods. While our approach employs in-context learning, commonly referred to as ``few-shot learning'', we opt to use ``zero-shot'' for consistency with previous literature on stance detection.} where the model is applied to new stance targets, unseen during training \cite{allaway-mckeown-2020-zero}.

\begin{figure}[t!]\centering
  \begin{tikzpicture}[font=\sffamily\small]
    \node [align=left,fill=blue!10, rounded corners=5pt,minimum size=2cm, text width=0.99\columnwidth] {
    \textbf{Target:} \textcolor{blue}{\textit{Hillary Clinton}}\hfill\textbf{Stance:} \textcolor{red}{\textit{Against}}\\
    
    \begin{minipage}{.05\textwidth}
        \includegraphics[width=1cm]{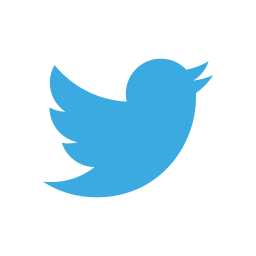}
    \end{minipage}%
    \begin{minipage}{.9\textwidth}
        \begin{quote}
        \textit{``She can't even manage her husband and she wants to be \textcolor{orange}{@POTUS}''}
        \end{quote}
    \end{minipage}
    
    \textbf{Background knowledge:}
    \begin{itemize}
        \item \textcolor{orange}{\textit{@POTUS}} is the Twitter handle for the president of the US.
        \item \textcolor{blue}{\textit{Hillary Clinton}} is the Democratic party nominee for president in the 2016 presidential election.
    \end{itemize}
    \textbf{Reasoning:} 
    \begin{itemize}
        \item \textbf{Premise}: Hillary Clinton is not qualified to be president because of her poor managing abilities.
        \item \textbf{Conclusion}: The author is against Hillary Clinton.
    \end{itemize}
    };
  \end{tikzpicture}
  \caption{\label{demo}
    An example of stance detection involves reasoning over background knowledge.
  }
\end{figure}

The concept of model generalization in zero-shot stance detection refers to a model's capacity to correctly identify the stance on new targets that it has not encountered before. The generalization ability depends on two factors. First, the model must capture background knowledge about the target. Second, the model needs to have general-purpose reasoning strategies over the context and background knowledge that it can apply to new targets. Consider the example in Figure~\ref{demo}. To make a prediction, a model should understand the context (\textit{``She wants to be @POTUS''}) and the background knowledge about the target (@POTUS is the Twitter handle for the president of the US, and Hillary Clinton is a presidential candidate). The model should reason that since the author is implying that Hillary Clinton is not qualified to be president because of her bad managing abilities, the stance is \texttt{against}.

Previous approaches to zero-shot stance detection typically involved fine-tuning a pre-trained language model (PLM) \cite{liu-etal-2021-enhancing,clark-etal-2021-integrating,he-etal-2022-infusing}. These supervised methods suffer from several drawbacks. First, these models may be learning features specific to the training targets, which negatively affects their ability to generalize to new targets \cite{kaushal-etal-2021-twt}. Second, even models that incorporate knowledge from external knowledge bases (KBs) may struggle from missing, sparse, or irrelevant knowledge, leading to subpar performance \cite{ma-etal-2019-towards}. Lastly, these models only output the predicted stance label without explaining the reason behind their prediction. The lack of transparency makes it challenging to understand the models' decision-making processes and address their errors.

To address these problems, in this paper, we present \textit{Stance Reasoner}, a framework for zero-shot stance detection on social media that leverages explicit reasoning over background knowledge to guide the model's inference about the document's stance on a target. To achieve this, \textit{Stance Reasoner} employs the \textit{in-context learning} approach \cite{brown2020language}. Unlike traditional methods that involve fine-tuning a PLM using a large training set, our approach involves providing the PLM with an optimized prompt. This approach, which avoids extensive training, enhances the model's capability to generalize effectively to new and unseen targets.

Specifically, our method utilizes a PLM as a source of world knowledge together with the \textit{chain-of-thought} (CoT) approach \cite{weiChainThoughtPrompting2022} to generate intermediate reasoning steps that lead to a label prediction. Therefore, our method not only predicts the stance label but also generates the underlying reasoning that supports its prediction. The ability to produce such explanations can help in understanding and debugging the models' decision-making processes. We demonstrate how our method can be used to detect annotation errors and ambiguous or otherwise difficult examples.

We evaluate \textit{Stance Reasoner} on three public Twitter stance detection datasets spanning a diverse range of targets. \textit{Stance Reasoner} outperforms all the baseline methods including the fully supervised state-of-the-art models. In addition, the results demonstrate that \textit{Stance Reasoner} can provide an interpretable and generalizable approach to zero-shot stance detection on social media. 

Our contributions are as follows:
\begin{itemize}
    \item We present \textit{Stance Reasoner}, a framework for zero-shot stance detection on social media that leverages explicit reasoning over background knowledge to guide the model's inference about the document's stance on a target and is based on the chain-of-thought (CoT) in-context learning.
    \item We analyze the impact of CoT on stance detection and show that the \textit{Stance Reasoner's} ability to reason using CoT depends on the diversity of reasoning strategies required for in-context examples.
    \item We demonstrate that our method outperforms the current state-of-the-art models on 3 Twitter datasets, including fully supervised models and it can better generalize across targets, while at the same time providing explicit and interpretable explanations for its predictions.
\end{itemize}

We make our code publicly available. \footnote{\url{https://github.com/maksym-taranukhin/stance_reasoner}}

\section{Methodology}
\label{sec:method}

In this work, we focus on \textit{zero-shot stance detection} \citep{allaway-mckeown-2020-zero}, which means the model is evaluated on a test set containing new targets that were never observed during training.

We propose \textit{Stance Reasoner}, a zero-shot stance detection approach. \textit{Stance Reasoner} uses CoT \cite{weiChainThoughtPrompting2022} to explicitly reason over background knowledge in order to guide the model's prediction regarding the document's stance on the target. In particular, we use a PLM and the CoT with the self-consistency approach to generate multiple intermediate reasoning steps that lead to the final prediction. Intermediate reasoning serves two purposes. First, it guides the model inference, and second, it provides a way to gain insights into the model's decision-making process.

We present the prompt (Sec~\ref{sec:prompt_formulation}), motivate the choice of in-context examples (Sec~\ref{sec:example_choice}), and describe the self-consistency approach that we use to further increase the model's accuracy (Sec~\ref{sec:self_consistency}).

\subsection{Prompt Formulation}
\label{sec:prompt_formulation}

At the core of our approach lies an optimized prompt that is used to induce the model to generate intermediate reasoning steps. As our experiments show, choosing the right prompt is key to the method's success. We design a prompt that consists of (i) the task description; and (ii) a set of examples augmented with the intermediate reasoning steps. Both are described below. 

\paragraph{Task description.} To provide the model with the best description of the task, we select the prompt with the highest likelihood according to a PLM. Following \newcite{gonenDemystifyingPromptsLanguage2022}, we first use a PLM to generate multiple paraphrases of manually defined seed task descriptions and then select the description that yields the lowest average perplexity on 100 random tweets. 

We use the following description of the stance detection task in the format of multiple-choice question answering, with the stance labels \texttt{against, favor, none} as answer candidates:

\begin{framed}
\ttfamily \footnotesize
\noindent Question: Consider the tweet in a conversation about the target, what could the tweet's point of view be towards the target?
\end{framed}

\paragraph{Examples.} To guide the model in generating intermediate reasoning steps, we provide a set of in-context examples, each with its respective reasoning and label. We define reasoning as an argument: the premise interprets the tweet, and supports the conclusion, which is the author's stance on the target. Therefore, each in-context example has the following format:
\begin{framed}
\ttfamily \small 
\noindent tweet: <tweet>\\
target: <stance target>\\
reasoning: <premise> -> <conclusion>\\
stance: <label>
\end{framed}

\subsection{Choice of In-Context Examples}
\label{sec:example_choice}

We argue that the context examples should cover a diverse set of reasoning strategies, both simple and more advanced, in order to help the model better generalize across documents and targets. Towards this end, we consider the reasoning strategies grouped based on two aspects: (1) target implicitness, i.e. whether the target is explicitly discussed in the document or whether it is implied, the latter requiring non-trivial reasoning strategy; and (2) the use of various rhetorical devices which might also require more complex reasoning. For example, whether the stance is expressed via sarcasm, jokes, aphorism, rhetorical question, etc. Ideally, we would like the prompt to cover examples from each group according to both aspects. In practice, exhaustively covering all possible reasoning strategies is not feasible in a short prompt. We thus limit the prompt to 6 examples, 2 examples for each label (\texttt{favor, against, none}). We include only examples with an implicit stance since they are more challenging for the model. Additionally, we include one example that uses sarcasm and another example that asks a rhetorical question. Finally, to comply with the zero-shot stance detection setup, the stance target of the in-context examples are distinct from the stance targets used in our experiments. Nonetheless, we observed that our prompt generalizes well across targets and datasets. 

\subsection{Self-Consistency}
\label{sec:self_consistency}

To further increase the model's prediction accuracy and its ability to generalize beyond the reasoning strategies present in the prompt, we employ the self-consistency approach \cite{wang2023selfconsistency}. Specifically, we generate multiple completions of the same data point and take a majority vote on the predicted labels as the final prediction. Other than increasing the accuracy of the model, self-consistency can also be used to spot examples that are inherently hard to predict, either due to the ambiguity naturally present in a tweet without additional context, or due to annotation errors. We define the model's prediction confidence as the ratio between the number of runs that predicted the majority label and the total number of runs. By considering confidence, we can recognize and eliminate unreliable predictions.

\section{Experiments}
We conducted experiments to evaluate the effectiveness of the proposed approach in modelling background knowledge and its impact on zero-shot stance detection. Below we provide the description of the datasets (Sec~\ref{sec:exp:dataset}), language models (Sec~\ref{sec:exp:models}), experimental setup (Sec~\ref{sec:exp:setup}), evaluation metrics (Sec~\ref{sec:exp:eval_metrics}), and baselines (Sec~\ref{sec:exp:baselines}).

\subsection{Datasets}
\label{sec:exp:dataset}
We conduct experiments on 3 Twitter datasets for stance detection, covering a wide range of domains. The SemEval-2016 Task 6a dataset \cite{mohammad-etal-2016-semeval} encompasses tweets pertaining to five targets across political, social, religious, and environmental domains. The WT-WT dataset \cite{conforti-etal-2020-will} focuses on tweets pertaining to corporate acquisition operations, along 5 targets, and with 4 labels (adding the \texttt{unrelated} label for tweets not discussing the target).  
Lastly, the COVID-19 Stance dataset \cite{glandt-etal-2021-stance} contains tweets discussing the coronavirus pandemic, featuring 4 targets within the public health domain.

We use the Twitter API\footnote{\url{https://developer.twitter.com/en/products/twitter-api}} to gather tweets from the WT-WT and COVID-19 datasets. Due to the limited accessibility of some tweets, the final dataset sizes are smaller than the originally collected. To form a testing split, we adopt different strategies for each dataset. For WT-WT, we randomly sample 100 data points for each target-label combination, yielding 2000 examples. Also, the \texttt{comment} and \texttt{unrelated} labels are merged into a single \texttt{none} label, ensuring that all datasets consist of 3 stance labels. For COVID-19, we fill in missing test data points using random samples from the training split with matching label-target combinations. Finally, we preprocess the SemEval-2016 Task 6a dataset to remove the \texttt{\#SemST} hashtag, which is not stance-indicative. 

\begin{table*}[!ht]
  \centering
  \small
  \begin{threeparttable}
    \begin{tabularx}{\textwidth}{lXXXXXr}
      \toprule
      \textbf{Model}                 & HC               & FM               & LA               & AT               & CC               & Avg              \\
      \midrule
      \multicolumn{7}{c}{\textit{Supervised Models}}                                                                                                   \\
      \textbf{BERT}\tnote{\dag}      & 49.6             & 41.9             & 44.8             & \underline{55.2} & 37.3             & 45.8             \\
      \textbf{TOAD}\tnote{\dag}      & 51.2             & \underline{54.1} & 46.2             & 46.1             & 30.9             & 45.7             \\
      \textbf{TGA Net}\tnote{\ddag}  & 49.3             & 46.6             & 45.2             & 52.7             & 36.6             & 46.1             \\
      \textbf{BERT-GCN}\tnote{\ddag} & 50.0             & 44.3             & 44.2             & 53.6             & 35.5             & 45.5             \\
      \textbf{JointCL}\tnote{\ddag}  & \underline{54.8} & 53.8             & \underline{49.5} & 54.5             & \underline{39.7} & \underline{50.5} \\
      \midrule
      \multicolumn{7}{c}{\textit{Unsupervised Models}\tnote{\S}}                                                                                       \\
      \textbf{Zero-Shot}                                                                                                                               \\
      \hspace{0.5cm}{Vicuna 13B}     & 55.6         & 61.4          &    45.3          &   7.2          &      62.5        &     46.4         \\
      \hspace{0.5cm}{LLaMA 65B}      & 26.6         & 31.8          & \underline{66.3} & 57.9 & 46.2 & 45.8              \\
      \textbf{Zero-Shot CoT}                                                                                                                           \\
      \hspace{0.5cm}{Vicuna 13B}     & \underline{67.9}         &  52.4           &  50.8           &   27.7           &   \underline{63.3}           &     52.4         \\
      \hspace{0.5cm}{LLaMA 65B}      & 25.1           & \underline{76.2}          & 47.2      & \underline{70.8}          & 47.2          & \underline{53.3}             \\
      \midrule
      \multicolumn{7}{c}{\textit{Few-Shot Models}\tnote{\S}}                                                                                           \\
      \textbf{Few-Shot}                                                                                                                                \\
      \hspace{0.5cm}{Vicuna 13B}     & 41.4         & 63.2         &    44.7          &  50.3            &    63.1          &   52.5           \\
      \hspace{0.5cm}{LLaMA 65B}      & 41.7 & 52.9 & 69.4 & 69.2 & 58.5 & 58.3             \\
      \textbf{Few-Shot CoT}                                                                                                                                \\
      \hspace{0.5cm}{Vicuna 13B}     & 72.0   & 65.3  &  66.1            &    52.2          &    65.7          &    64.3          \\
      \hspace{0.5cm}{LLaMA 65B}      & 69.1	& \underline{67.7}	& \underline{72.9} & 	\underline{71.2}	& 61.8 & 	\underline{68.5}             \\
      \textbf{Stance Reasoner (ours)}                                                                                                      \\
      \hspace{0.5cm}{Vicuna 13B}     & \textbf{74.4}       &   66.8           &    67.6          &  53.3            & \underline{67.4}             &  65.9            \\
      \hspace{0.5cm}{LLaMA 65B}      & \underline{73.7} & \textbf{76.2} & \textbf{79.4} & \textbf{75.7} & \textbf{68.1} & \textbf{72.6}            \\
      \bottomrule
    \end{tabularx}
    \begin{tablenotes}
      \item[\dag] Results reported by \cite{allaway-etal-2021-adversarial}
      \item[\ddag] Results reported by~\cite{liang-etal-2022-jointcl}
      \item[\S] Our implementation
    \end{tablenotes}
  \end{threeparttable}
  \caption{
    \label{tab:results}
    Experimental results on the SemEval 2016 task 6a dataset. We report macro $F_1$ scores for each target in the test split, namely,
    HC - \textit{Hillary Clinton},
    FM - \textit{Feminist Movement},
    LA - \textit{Legalization of Abortion},
    AT - \textit{Atheism},
    CC - \textit{Climate Change is a Real Concern}
    . The best results are highlighted in bold. The second-best results for each group of the baseline models are highlighted with underlining.
  }
\end{table*}

\begin{table}[!ht]
  \small
  \begin{threeparttable}
    \begin{tabularx}{\columnwidth}{lcccl}
      \toprule
      \textbf{Model}             & SemEval & Covid-19 & WT-WT     \\
      \midrule
      \multicolumn{4}{c}{\textit{Unsupervised Models}\tnote{\S}} \\
      \textbf{Zero-Shot}                                     \\
      \hspace{0.5cm}{Vicuna 13B} & 46.6    & 48.5     & 65.7 \\
      \hspace{0.5cm}{LLaMA 65B}  & 45.8    & 31.8     & \underline{66.3} \\
      \textbf{Zero-Shot CoT}                                 \\
      \hspace{0.5cm}{Vicuna 13B} & 52.4   & 53.1     &  35.1 \\
      \hspace{0.5cm}{LLaMA 65B}  & \underline{53.3}   & \underline{55.8}    & 41.4   \\
      \midrule
      \multicolumn{4}{c}{\textit{Few-Shot Models}\tnote{\S}} \\
      \textbf{Few-Shot}                                      \\
      \hspace{0.5cm}{Vicuna 13B} & 52.5    &  51.1   &   64.7   \\
      \hspace{0.5cm}{LLaMA 65B}  & 58.3    & 52.9     & 69.4 \\
      \textbf{Few-Shot CoT}                                  \\
      \hspace{0.5cm}{Vicuna 13B} &  64.3    &  74.9   &  69.8    \\
      \hspace{0.5cm}{LLaMA 65B}  & \underline{68.5}    & \underline{75.5}     & \underline{73.7} \\
      \textbf{Stance Reasoner}                        \\
      \hspace{0.5cm}{Vicuna 13B} &  65.9       &  74.6        &  73.6    \\
      \hspace{0.5cm}{LLaMA 65B}  & \textbf{72.6}    & \textbf{76.2}     & \textbf{78.3} \\
      \bottomrule   \\
    \end{tabularx}
    \begin{tablenotes}
      \item[\S] Our implementation
    \end{tablenotes}
  \end{threeparttable}
  \caption{
    \label{tab:prompt_generalization}
    Generalization results across three datasets. We report average macro $F_1$ scores for all targets in the test split. The best results are highlighted in bold. The second-best results for each group of the baseline models are highlighted with underlining.
  }
\end{table}

\subsection{Models}
\label{sec:exp:models}
We evaluate our approach on a range of open-source auto-regressive language models due to their strong in-context learning abilities~\cite{pmlr-v162-wang22u}. Specifically, we use LLaMA 65B \cite{touvron2023llama} - an open-source model trained on publicly available datasets, and the Vicuna (13B) model \cite{vicuna2023} - the LLaMA models that are finetuned to follow instructions with reinforcement learning \cite{NEURIPS2022_b1efde53}.

\subsection{Setup}
\label{sec:exp:setup}
All models are used for inference only. We utilize the HuggingFace Transformers library \cite{wolf-etal-2020-transformers} to load the LLaMA models in half-precision and run them using 4 A100 40GB GPUs. In all experiments, the maximum sequence length is set to 256 and the temperature is set to 0. When we sample multiple completions, the number of samples per tweet is set to 5 and the temperature is set to 0.7. We generate 50 paraphrases of each seed description using LLaMA 65B.

\subsection{Evaluation Metric}
\label{sec:exp:eval_metrics}
Following prior work, we report the macro-averaged $F_1$ score across the \texttt{against} and \texttt{favor} labels for each of the targets in the dataset.

\subsection{Baselines}
\label{sec:exp:baselines}
We compare our approach to several baselines depending on the dataset. To simulate the zero-shot stance detection setting, we follow the leave-one-target-out evaluation setup. That is when the model is trained on all but one target which is held out for evaluation. However, since our method does not make use of the dataset's training split, we just measure the performance of each target in the test set individually.

\subsubsection{Supervised Baselines}

The supervised approaches are evaluated on SemEval-2016 Task 6a only.

\paragraph{BERT-base \cite{allaway-etal-2021-adversarial}.} A vanilla BERT-base model with a classification head. The input is represented as \texttt{[CLS]<tweet>[SEP]<target>[SEP]}.

\paragraph{TGA Net \cite{allaway-mckeown-2020-zero}.} The model uses unsupervised clustering of BERT-embeddings together with attention to improve performance on new targets.

\paragraph{TOAD \cite{allaway-etal-2021-adversarial}.} A BiLSTM model that uses adversarial learning to produce topic-invariant representations for better generalization to new targets.

\paragraph{BERT-GCN \cite{liu-etal-2021-enhancing}.} A knowledge-infused model that uses conventional GCN to embed the nodes of a sub-graph consisting of entities extracted from ConceptNet \cite{speer2017conceptnet}.

\paragraph{JoinCL \cite{liang-etal-2022-jointcl}.} A join contrastive learning framework for zero-shot stance detection that combines stance contrastive learning and target-aware prototypical graph contrastive learning.

\subsubsection{Unsupervised Baselines}

The unsupervised approaches are evaluated on all datasets.

\paragraph{Zero-Shot \cite{kojima2022large}.} Unsupervised zero-shot stance detection via multiple-choice question answering. We provide a task description and a test example in the prompt and let the model generate the answer with greedy decoding. Similarly to our model, we chose the prompt that yielded the lowest average perplexity on 100 random examples. The prompt optimization is performed separately for each model and size. 

\paragraph{Zero-Shot CoT \cite{kojima2022large}.} We use the prompt from the zero-shot setup and engage the model in reasoning with zero-shot CoT using a two-step approach: 1) \textit{reasoning generation} via appending  {\ttfamily Let's think step by step} to the input, followed by 2) \textit{answer prediction} by concatenating the generated reasoning to the input together with an answer trigger {\ttfamily Therefore, the answer is}. We use greedy decoding and parse the second step output to extract the prediction. 

\subsubsection{Few-Shot Baselines}

\paragraph{Few-Shot \cite{brown2020language}.} We modify the prompt from the zero-shot setup to include 6 in-context examples, two for each label. These examples have explicit and implicit stances towards the targets that are not present in the dataset. 

\paragraph{Few-Shot CoT \cite{weiChainThoughtPrompting2022}.} We use the same prompt and in-context examples as in the few-shot baseline, but augment the examples with manually-written reasoning chains. 

\begin{table*}[!ht]
\centering
\small
\begin{tabularx}{\textwidth}{lXXXXXr}
\toprule
\textbf{Model} & AT & CC & FM& HC & LA & Avg \\
\midrule
\multicolumn{7}{l}{\textbf{Homogeneous CoT Prompt}} \\
\hspace{0.5cm}\textit{without Self-Consistency}   & 71.0 & 63.8 & 65.6 & 69.1 & 64.7 & 66.9 \\
\hspace{0.5cm}\textit{with Self-Consistency}      & 71.2 & 65.4 & 64.1 & \textbf{76.4} & 64.8 & 68.4 \\
\multicolumn{7}{l}{\textbf{Stance Reasoner}} \\
\hspace{0.5cm}\textit{without Self-Consistency}   & \underline{72.2} & \underline{65.5} & \underline{75.1} & \underline{75.7} & \underline{66.8} & \underline{70.6} \\
\hspace{0.5cm}\textit{with Self-Consistency}      & \textbf{73.7} & \textbf{66.2} & \textbf{79.4} & \underline{75.7} & \textbf{68.1} & \textbf{72.6} \\
\bottomrule
\end{tabularx}

\caption{
    \label{tab:ablation_study}
    Ablation study comparing the performance of Stance Reasoner to a less diverse, homogeneous CoT prompt. We report macro $F_1$ scores for each target in the SemEval 2016 task 6a test split, namely, 
    HC - \textit{Hillary Clinton},
    FM - \textit{Feminist Movement},
    LA - \textit{Legalization of Abortion},
    AT - \textit{Atheism},
    CC - \textit{Climate Change is a Real Concern}. The best results are highlighted in bold. The second-best results are highlighted with underlining.
}
\end{table*}

\section{Results}
\label{sec:results}
Table~\ref{tab:results} displays the performance of \textit{Stance Reasoner} and the baselines on the SemEval 2016 task 6a test set, in terms of the macro-$F_1$ score for each target. \textit{Stance Reasoner} outperforms not only all the unsupervised baselines but also the supervised baselines, including a knowledge-infused model BERT-GCN, and across all targets---despite seeing only 6 examples. In fact, supervision seems to be detrimental in the leave-one-target-out setup, and our few-shot approach surpasses the best-supervised method by between 20 and 30 $F_1$ points.

While the supervised models differ in their base language model, we can see that even compared to the few-shot model, \textit{Stance Reasoner} achieves better performance across targets, with very large gaps on some targets (e.g., 32 points on atheism). In general, few-shot methods performed better than zero-shot methods, and adding CoT to zero-shot degraded the performance. We attribute this to the fact that the few-shot approach has access to the set of supporting examples compared to the zero-shot CoT approach.

Overall, our method achieved the best average $F_1$ score of 72.6. The results suggest that the proposed approach is able to effectively use reasoning to infer the correct stance of a document on a target. This leave-one-target-out setup shows the method's ability to generalize its reasoning strategies across targets. We attribute this to the fact that the prompt contains diverse reasoning strategies that the model can learn to employ, and to the self-consistency strategy that is more robust to the randomness of the decoding strategy.

\begin{figure}[!t]
\centering
\includegraphics[width=\columnwidth]{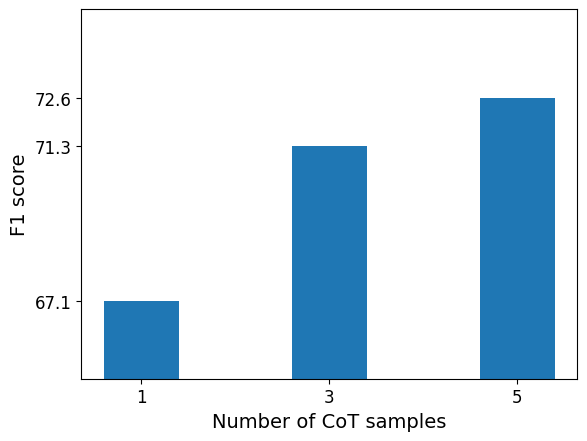}
\caption{\label{fig:num_samples_in_self_consistency}
The impact of the number of sampled reasonings on the performance of Stance Reasoner. The performance increases with the number of samples.}
\end{figure}

\begin{table*}[!t]
  \centering
  \small
  \begin{tabularx}{.93\textwidth}{lXllXr}
    \toprule

    \textbf{\#}                                                                              &
    \textbf{Example}                                                                         &
    \textbf{Gold}                                                                            &
    \textbf{Pred}                                                                            &
    \textbf{Reasoning}                                                                       &
    \textbf{Conf.}

    \\
    \midrule
    \small{\textcircled{\scriptsize 1}}                                                      &
    \textbf{Tweet:} \textit{It's most exciting to witness a major development! @urgenda}\newline
    \textbf{Target:} \textit{Climate Change is a Real Concern}                               &
    \texttt{\footnotesize none}                                                                &
    \texttt{\footnotesize favor}                                                               &
    \textbf{Premise:} \textit{the author is excited about the major development of an organization that tackles climate change}\newline
    \textbf{Conclusion:} \textit{the author is in favor of climate change is a real concern} &
    1.0

    \\
    \midrule
    \small{\textcircled{\scriptsize 2}}                                                      &
    \textbf{Tweet:} \textit{One thing I learned from my job: doors to opportunity cover fee that only the privileged can afford. \#privilege \#truth}\newline
    \textbf{Target:} \textit{Feminist Movement}                                              &
    \texttt{\footnotesize against}                                                             &
    \texttt{\footnotesize none}                                                                &
    \textbf{Premise:} \textit{the author is in favor of truth}\newline
    \textbf{Conclusion:} \textit{the author is neutral towards the feminist movement}        &
    0.4

    \\
    \midrule
    \small{\textcircled{\scriptsize 3}}                                                      &
    \textbf{Tweet:} \textit{@cbrangel so, you support the choice of wether or not you'd like to kill someone? Would you kill a born baby?}\newline
    \textbf{Target:} \textit{Legalization of Abortion}                                       &
    \texttt{\footnotesize against}                                                             &
    \texttt{\footnotesize against}                                                             &
    \textbf{Premise:} \textit{the author is against the idea of people choosing to kill other people}\newline
    \textbf{Conclusion:} \textit{the author is against the legalization of abortion.}        &
    0.4
    \\
    \bottomrule
  \end{tabularx}
  \caption{
    \label{tab:qual_analysis}
    Examples of tweets along with their gold label (\textbf{Gold}), and LLaMA 65B model-predicted label (\textbf{Pred}), reasoning chain (\textbf{Reasoning}), and self-consistency confidence as described in Sec~\ref{sec:self_consistency} (\textbf{Conf.}). We show examples of: (1) annotation error, (2) ambiguous context, and (3) rhetorical devices.
  }
\end{table*}

\subsection{Ablation Tests}
\label{sec:ablations}

\paragraph{Impact of Diverse Examples.} We analyze the impact of choosing examples with diverse reasoning strategies on the final performance of our model. To that end, we compare the performance of \textit{Stance Reasoner} on the SemEval 2016 task 6a test set with the performance of an identical model that differs only by including only examples with explicit stances and without rhetorical devices (homogeneous in Table~\ref{tab:ablation_study}). 

\textit{Stance Reasoner} outperforms the homogeneous CoT prompt by a large margin. Furthermore, our approach outperforms the homogeneous CoT prompt even when the homogeneous model uses self-consistency and the diverse prompt doesn't. We conclude that including diverse reasoning strategies in the CoT prompt is beneficial for few-shot stance detection models.

\paragraph{Impact of Self-Consistency.} We also evaluate the impact of the self-consistency strategy on the model performance. Table~\ref{tab:ablation_study} shows that when we remove self-consistency (i.e. only generate one output), we get a substantial drop in performance. Fig~\ref{fig:num_samples_in_self_consistency} further shows the performance of our approach with self-consistency with a different number of sampled completions. The model performs better as we increase the number of samples. Self-consistency increases the model's robustness to noisy generations, leading to better performance.

\paragraph{Prompt generalization.} We evaluate the generalization ability of our prompts by testing them on 3 Twitter datasets from different domains. Table~\ref{tab:prompt_generalization} shows that our method based on the CoT prompt achieves state-of-the-art performance on all the datasets. This suggests that the CoT prompt is not domain-specific and can generalize to other domains without model fine-tuning.

\paragraph{Model size.} We also analyze the effect of the model size on the performance of our approach. Table~\ref{tab:results} and Table~\ref{tab:prompt_generalization} show that the larger LLaMA 65B model consistently achieves better performance. However, the performance of Vacuna 13B is also competitive especially considering its size is 5 times smaller than LLaMA 65B. We hypothesize Vacuna's performance can be attributed to the ability of the model to follow instructions.


\section{Qualitative Analysis} 
\label{sec:analysis:confidence}

Table~\ref{tab:qual_analysis} shows the reasoning chains and labels predicted by \textit{Stance Reasoner} for a few example tweets from the SemEval test set. We show how the model's confidence score (Sec~\ref{sec:self_consistency}) can be used to detect annotation errors, and ambiguous or difficult contexts.

\begin{enumerate}[label=\large\protect\textcircled{\small\arabic*}]

  \item \textbf{Annotation Error.} The annotators marked this tweet as neutral instead of in favor of the stance ``Climate change is a real concern''. This is likely due to the absence of background knowledge that Urgenda is a pro-environment nonprofit organization. \textit{Stance Reasoner} predicted the correct label and reasoning, thanks to its access to world knowledge from a PLM. Moreover, all self-consistency samples predicted the same label, as indicated by the confidence score. By employing this approach and manually inspecting tweets and corresponding reasoning we additionally found a total of 88 annotation errors in the dataset.

  \item \textbf{Ambiguous Context.} We also observe that some tweets are ambiguous without additional context. For example, whether the author of the tweet is discussing male privilege specifically is unclear from the context. In those cases, the model's confidence tends to be lower. Despite the low confidence and the incorrect label predicted by the model for this example, the generated reasoning is logically sound.

  \item \textbf{Rhetorical Device.} Finally, we observed low confidence predictions for tweets containing rhetorical devices such as rhetorical questions.
\end{enumerate}

The results suggest that the proposed approach effectively uses reasoning over background knowledge to predict the correct label, and can even help identify annotation errors. Furthermore, the model is able to provide explicit and interpretable explanations for its predictions.

\section{Prior Work}
\label{sec:bg:related_work}

\paragraph{Target-specific stance detection.}
In this setup, the model is trained and evaluated on the same set of stance targets.  
Previous work on \emph{target-specific} stance detection primarily employed additional knowledge sources of information to alleviate the problem of an implicit stance target, such as in Fig.~\ref{demo}   \cite{xuOpinionAwareKnowledgeEmbedding2019,duCommonsenseKnowledgeEnhanced2020,sunStanceDetectionKnowledge2021}. Typically, a subgraph of relevant knowledge pertaining to the words in the stance document and target is extracted from KBs such as DBPedia \cite{auer2007dbpedia} or ConceptNet \cite{speer2017conceptnet}, and incorporated into the stance detection model. 
\newcite{zhangKnowledgeEnhancedTargetAware2021} selected relevant concepts from multiple knowledge bases by measuring cosine-similarity between the BERT-embeddings of each concept and potential concepts (n-grams) in the document. \newcite{clark-etal-2021-integrating} employed Wikidata \cite{vrandevcic2014wikidata} to provide definition concepts to a language model as raw text.

\paragraph{Cross-target and zero-shot stance detection.} 
Cross-target stance detection aims to predict stances for new targets related to the train targets, while zero- and few-shot stance detection aims to predict stance for entirely unrelated targets with no or little training data. In both setups, external knowledge may be used to help uncover implicit targets and generalize to new ones. Prior work incorporated relevant Wikipedia articles \cite{hanawaStanceDetectionAttending2019a}, semantic and emotion lexicons \cite{zhang-etal-2020-enhancing-cross}, and knowledge from ConceptNet \cite{liu-etal-2021-enhancing}. However, as shown recently, such extracted knowledge is not consistently helpful for the model to make predictions \cite{chan2021salkg, raman2021learning}. In addition, \newcite{he-etal-2022-infusing} noted that retrieving relevant Wikipedia articles sometimes required manual work, and some targets lacked Wikipedia pages entirely \cite{he-etal-2022-infusing}.

\paragraph{Prompt-based and in-context learning approaches.} Since all these methods are fully supervised, they tend to overfit the target-specific features and fall short when predicting the stance of new targets. In this work, we propose an in-context learning method that requires only a small number of labeled examples, preserving the model's generality which contributes to higher performance in zero-shot stance detection.

Recently, with the wide adoption of large LMs and in-context learning, there has been parallel work that also explored the use of prompts to perform stance detection. \newcite{zhangInvestigatingChainofthoughtChatGPT2023} showed that CoT prompting with ChatGPT can outperform zero- and few-shot supervised learning approaches. However, the work didn't study the impact of the prompt selection on the performance, used a subset of the SemEval stance dataset targets and employed a closed-source model that limits the reproducibility of the work.

Our work is closely related to techniques for automatic CoT prompt construction, such as Auto-CoT~\cite{zhang2023automatic}, but stands out in several key aspects. Firstly, our approach employs a fixed prompt structure, which enables cross-dataset generalization and is specifically tailored for stance detection. Secondly, we not only assess the performance but also conduct a detailed analysis of the LM's reasoning abilities within the realm of stance detection. This is in contrast to methods like Auto-CoT, which primarily focus on enhancing performance without thoroughly examining the LM's reasoning structure and validity.


\paragraph{Chain-of-Thought Prompting} \textit{Chain-of-thought} \cite[CoT;][]{weiChainThoughtPrompting2022} is an in-context learning approach that uses a language model to generate a sequence of intermediate reasoning steps that lead to the final prediction. \emph{Few-shot} CoT builds a prompt to the model that consists of an optional task description $T$ and a set of $M$ examples. Each example $\{(x_i, r_i, y_i)\}_{i=1}^{M}$ consists of the input $x_i$, and the intermediate reasoning steps $r_i$ that lead to label $y_i$. To obtain a prediction, the prompt is concatenated with a new data point $x'_i$, for which the model needs to generate the reasoning steps and the predicted label. \emph{Zero-shot} CoT \citep{kojima2022large} excludes the set of $M$ examples from the model's prompt and instead appends the text ``Let's think step by step'' to encourage the language model to generate the reasoning $r$ in an unsupervised way. In a subsequent step, ``Therefore, the answer is'' is appended to the reasoning to predict the label. A further improvement, known as CoT with \emph{self-consistency} can be achieved by sampling multiple completions for the same data point and taking a majority vote among them to obtain the final prediction \cite{wang2023selfconsistency}.

Building on this foundation, our stance reasoner improves the traditional CoT method by including examples that demonstrate various reasoning strategies designed specifically for detecting stances in texts. It also uses a clear reasoning format that moves from the starting point to the conclusion, making it better suited for stance detection. These changes highlight what sets our work apart from regular CoT methods and emphasize its effectiveness in accurately analyzing and understanding social media texts with reasoning.

\section{Conclusion}
We presented Stance Reasoner, a zero-shot stance detection model. Stance Reasoner generates explicit reasoning over background knowledge to predict the stance of a given tweet regarding a target. Our empirical results show that Stance Reasoner outperforms the current state-of-the-art models on a Twitter dataset, including fully supervised models, and that it can better generalize to new targets, domains and datasets. We also presented a qualitative analysis of the model's performance, showing that it can accurately identify annotation errors and generate interpretable explanations for its predictions.

In the future, we plan to develop a model that can better handle tweets including rhetorical devices such as sarcasm and rhetorical questions, as well as tweets containing quotations. We will also further instigate the types of knowledge that are missing from language models and how to supplement the model with such knowledge. Finally, we also will aim to extend our method's application beyond concise texts like tweets to longer formats such as opinion pieces or blog posts. Given the more scattered and implicit presentation of information in these longer texts, adapting our method to accommodate the extended reasoning chains will pose a significant challenge.

\section{Limitations}

\paragraph{Language Models.} The proposed approach relies on the knowledge encoded in a PLM. We expect the model's performance and generalization ability to degrade if tested on brand-new topics on which the PLM doesn't contain information. In addition, Stance Reasoner is most effective with larger models, which might be prohibitively expensive to run and geographically limited to some regions.

\paragraph{Social Media Text.} We tested Stance Reasoner on datasets in the social media domain, where texts tend to be short and noisy. Although our approach is not designed specifically for this domain, the question of whether it can generalize to other domains or longer texts (e.g., news articles) is an interesting future research direction.

\paragraph{CoT Faithfulness and Task Definition.} While CoT generates the intermediate reasoning steps leading to the prediction, there is no guarantee that the prediction causally depends on the reasoning steps \cite{creswell2023selectioninference}. While we were able to manually verify the correctness of a sample of the reasoning chains, we also note that judging some of the examples required substantial efforts and sometimes extra context. For example, a tweet such as ``When women spend too much time out of the kitchen they get over opinionated and think they know everything \#feminist'' seems at face value to be against the feminist movement; however it may be interpreted in favor of it if it was written by a feminist user, as a sarcastic response to a misogynistic tweet. We thus advocate for future research to adapt the stance detection task from a classification task to a more flexible format where models can generate multiple interpretations along with their reasoning.

\section{Ethical Considerations}

We proposed a tool for automated stance detection on social media. As with any automated tool, it has the potential of being used in unintended ways and amplifying existing social issues such as political polarization. Thus, while the proposed approach can be used for various positive applications, such as identifying and addressing fake news, it is important to consider ethical implications and potential harms to individuals and society when deploying the proposed approach in real-world applications.

The datasets used in this paper are publicly available for research purposes on the owners' website.

\section*{Acknowledgements}
We would like to express our sincere gratitude to Compute Canada for providing the computational resources that were instrumental in conducting the experiments and analysis presented in this paper.

Furthermore, we would like to acknowledge the invaluable contributions of Diversio. The unique insights gained from this partnership greatly enriched the quality and impact of our research.

Finally, Vered's research is supported by the Vector Institute for AI, the CIFAR AI Chair program, and NSERC.

\nocite{*}
\section{Bibliographical References}\label{sec:reference}

\bibliography{lrec-coling2024-example}
\bibliographystyle{lrec-coling2024-natbib}


\newpage

\appendix
\onecolumn
\section{Dataset Details}
\label{sec:appendix:dataset}


\begin{table}[ht]
  \centering
  \small
  \begin{tabularx}{\textwidth}{p{4.7cm}*{8}{>{\centering\arraybackslash}X}}
    \toprule
    \multirow{2}[2]{*}{\bf{Target}}       & \multicolumn{4}{c}{train split counts} & \multicolumn{4}{c}{test split counts}                                                 \\
    \cmidrule(lr){2-5} \cmidrule(lr){6-9}
                                          & against                                & favor                                 & none & total & against & favor & none & total \\
    \midrule
    \bf{Atheism}                          & 304                                    & 92                                    & 117  & 513   & 160     & 32    & 28   & 220   \\
    \bf{Climate Change} & 15                                     & 212                                   & 168  & 395   & 11      & 123   & 35   & 169   \\
    \bf{Feminist Movement}                & 328                                    & 210                                   & 126  & 664   & 183     & 58    & 44   & 285   \\
    \bf{Hillary Clinton}                  & 393                                    & 118                                   & 178  & 689   & 172     & 45    & 78   & 295   \\
    \bf{Legalization of Abortion}         & 355                                    & 121                                   & 177  & 653   & 189     & 46    & 45   & 280   \\
    \midrule
    \bf{All Targets}                      & 1395                                   & 753                                   & 766  & 2914  & 715     & 304   & 230  & 1249  \\
    \bottomrule
  \end{tabularx}
  \caption{The SemEval 2016 Task 6a dataset~\cite{mohammad-etal-2016-semeval} count statistics. The splits are provided by the dataset authors.}
\end{table}

\begin{table}[ht]
  \centering
  \small
  \begin{tabularx}{\textwidth}{p{3.7cm}*{5}{>{\centering\arraybackslash}X}}
    \toprule
    \textbf{Target}                    & refute      & support     & comment       & unrelated     & total         \\
    \midrule
    \bf{Aetna $\to$ Humana}            & 717 (1106)  & 728 (1038)  & 1937 (2804)   & 1925 (2949)   & 5307 (7897)   \\
    \bf{Anthem $\to$ Cigna}            & 1286 (1969) & 682 (970)   & 2068 (3098)   & 3293 (5007)   & 7329 (11622)  \\
    \bf{CVS Health $\to$ Aetna}        & 294 (518)   & 1323 (2469) & 3016 (5520)   & 1618 (3115)   & 6251 (11622)  \\
    \bf{Cigna $\to$ Express Scripts}   & 140 (253)   & 408 (773)   & 506 (947)     & 306 (554)     & 1360 (2527)   \\
    \bf{Disney $\to$ 21st Century Fox} & 217 (378)   & 797 (1413)  & 4568 (8495)   & 4019 (7908)   & 9601 (18194)  \\
    \midrule
    \bf{All Targets}                   & 2654 (4224) & 3938 (6663) & 12095 (20864) & 11161 (19533) & 29848 (51284) \\
    \bottomrule
  \end{tabularx}
  \caption{The WT-WT dataset \cite{conforti-etal-2020-will} count statistics. The counts represent the number of data points accessible via Twitter API. The original counts provided by the dataset authors are enclosed in parentheses.}
\end{table}

\begin{table}[ht]
  \centering
  \small
  \begin{tabularx}{\textwidth}{p{4cm}*{12}{>{\centering\arraybackslash}X}}
    \toprule
    \multirow{2}[2]{*}{\bf{Target}} & \multicolumn{4}{c}{train split counts} & \multicolumn{4}{c}{val split counts} & \multicolumn{4}{c}{test split counts} \\
    \cmidrule(lr){2-5} \cmidrule(lr){6-9} \cmidrule(lr){10-13}
                                    & again.                                & favor                                & none                                  & total       & again.   & favor     & none      & total     & again.   & favor     & none      & total     \\
    \midrule
    \bf{Anthony S. Fauci, M.D.}     & 211 (480)                              & 273 (388)                            & 312 (596)                             & 796 (1464)  & 28 (65)   & 39 (52)   & 39 (83)   & 106 (200) & 27 (65)   & 43 (52)   & 38 (83)   & 108 (200) \\
    \bf{Keeping Schools Closed}     & 98 (166)                               & 259 (409)                            & 135 (215)                             & 492 (790)   & 26 (42)   & 69 (103)  & 41 (55)   & 136 (200) & 19 (42)   & 64 (103)  & 37 (55)   & 120 (200) \\
    \bf{Stay at Home Orders}        & 137 (284)                              & 104 (136)                            & 417 (552)                             & 658 (972)   & 30 (58)   & 18 (27)   & 89 (115)  & 137 (200) & 31 (58)   & 18 (27)   & 92 (115)  & 141 (200) \\
    \bf{Wearing a Face Mask}        & 206 (512)                              & 366 (531)                            & 170 (264)                             & 742 (1307)  & 31 (78)   & 52 (81)   & 22 (41)   & 105 (200) & 35 (78)   & 52 (81)   & 24 (41)   & 111 (200) \\
    \midrule
    \bf{All targets}                & 652 (1442)                             & 1002 (1464)                          & 1034 (1627)                           & 2688 (4533) & 115 (243) & 178 (263) & 191 (294) & 484 (800) & 112 (243) & 177 (263) & 191 (294) & 480 (800) \\
    \bottomrule
  \end{tabularx}
  \caption{The COVID 19 Stance dataset~\cite{glandt-etal-2021-stance} count statistics. The counts represent the number of data points accessible via Twitter API. The original counts provided by the dataset authors are enclosed in parentheses. The splits are provided by the dataset authors.}
\end{table}
\section{Model Details}

We used the following model checkpoints from HuggingFace Hub in our experiments:

\begin{itemize}
    \item decapoda-research/llama-65b-hf
    \item TheBloke/vicuna-13B-1.1-HF
\end{itemize}

\newpage
\section{Prompts}
\label{sec:appendix:prompts}

\subsection{Prompt Generation}
\label{sec:appendix:prompts:prompt_generation}
We manually define the following seed task descriptions:
\begin{itemize}
    \item \it{What is the tweet's stance on ``\{target\}''?}
    \item \it{In the context of a discussion about ``\{target\}'', what could be the tweet's stance on ``\{target\}''?}
\end{itemize}

We used the following meta-prompt to paraphrase the seed task descriptions with LLaMA-65B and sampling temperature = 0.7:
\begin{framed}
  \ttfamily \small \noindent
    Write 50 diverse paraphrases for the following sentence: <seed-prompt>.
    Paraphrases:
\end{framed}

The perplexity of a task description is measured on random 100 examples sampled from the train split of the SemEval-2016 dataset and format with the below prompt. Note: we did not use labels in the prompt therefore this procedure is unsupervised.
\begin{framed}
  \ttfamily \small \noindent
  Question: \{task description\}\\
  The options are:\\
  - against\\
  - favor\\
  - none\\
  tweet: <\{text\}>\\
  Answer:
\end{framed}

Utilizing this approach we found that the best prompt is the same among the models. We explain this observation due the fact Vicuna is a finetuned LLaMA model and therefore shares the same pre-training data.

\subsection{Zero-Shot Prompt}
\label{sec:appendix:prompts:zero-shot}
\begin{framed}
  \ttfamily \small \noindent
  Question: In a conversation about ``\{target\}'', what could the tweet's point of view be towards ``\{target\}''?\\
  The options are:\\
  - against\\
  - favor\\
  - none\\
  tweet: <\{text\}>\\
  Answer: The tweet could be
\end{framed}

\subsection{Zero-Shot CoT Details}
\label{sec:appendix:prompts:zero-shot-cot}

The stance detection prompt used in zero-shot CoT:

\begin{framed}
  \ttfamily \small \noindent
  Question: In a conversation about ``\{target\}'', what could the tweet's point of view be towards ``\{target\}''?\\
  The options are:\\
  1. against\\
  2. favor\\
  3. none\\
  tweet: <\{text\}>\\
  Answer: Let's think step by step. \{CoT\}. Therefore, the answer is
\end{framed}

We used the following regular expression to find the first occurrence of an option number concatenated with a dot and take the corresponding option word as the final prediction: \texttt{(1|2|3)}.

\newpage
\subsection{Few-Shot Prompt and Few-Shot CoT Prompt}
\label{sec:appendix:prompts:few-shot}
\label{sec:appendix:prompts:few-shot-cot}

The Stance Reasoner few-shot prompt with reasoning chains is highlighted in blue.
\begin{framed}
  \ttfamily \small \noindent
    Question: Consider the tweet in a conversation about the target, what could the tweet's point of view be towards the target?\newline
  The options are: \newline
  - against \newline
  - favor \newline
  - none \newline
  \newline
  tweet: <I'm sick of celebrities who think being a well known actor makes them an authority on anything else. \#robertredford \#UN> \newline
  target: Liberal Values \newline
  {\color{blue}reasoning: the author is implying that celebrities should not be seen as authorities on political issues, which is often associated with liberal values such as Robert Redford who is a climate change activist -> the author is against liberal values} \newline
  stance: against \newline
  \newline
  tweet: <I believe in a world where people are free to move and choose where they want to live>\newline
  target: Immigration\newline
  {\color{blue}reasoning: the author is expressing a belief in a world with more freedom of movement -> the author is in favor of immigration.}\newline
  stance: favor\newline
  \newline
  tweet: <I love the way the sun sets every day. \#Nature \#Beauty>\newline
  target: Taxes\newline
  {\color{blue}reasoning: the author is in favor of nature and beauty -> the author is neutral towards taxes}\newline
  stance: none\newline
  \newline
  tweet: <If a woman chooses to pursue a career instead of staying at home, is she any less of a mother?>\newline
  target: Conservative Party\newline
  {\color{blue}reasoning: the author is questioning traditional gender roles, which are often supported by the conservative party -> the author is against the conservative party}\newline
  stance: against\newline
  \newline
  tweet: <We need to make sure that mentally unstable people can't become killers \#protect \#US>\newline
  target: Gun Control\newline 
  {\color{blue}reasoning: the author is advocating for measures to prevent mentally unstable people from accessing guns -> the author is in favor of gun control.}\newline
  stance: favor\newline
  \newline
  tweet: <There is no shortcut to success, there's only hard work and dedication \#Success \#SuccessMantra>\newline
  target: Open Borders\newline
  {\color{blue}reasoning: the author is in favor of hard work and dedication -> the author is neutral towards open borders}\newline
  stance: none
\end{framed}

\newpage
\subsection{Homogeneous Few-Shot Prompt and Few-Shot CoT Prompt}
\label{sec:appendix:prompts:homo-few-shot}
\label{sec:appendix:prompts:homo-few-shot-cot}

Homogeneous few-shot prompt with reasoning chains is highlighted in blue.

\begin{framed}
  \ttfamily \small \noindent
  Question: Consider the tweet in a conversation about the target, what could the tweet's point of view be towards the target?\newline
  The options are: \newline
  - against \newline
  - favor \newline
  - none \newline
  \newline
    tweet: <RT @MyDailyMeat: Real food is MEAT, not vegetables. Humans were built to eat meat, not vegan diets. \#meatlover \#notvegan \#realfood> \newline
    target: Veganism \newline
    {\color{blue}reasoning: the author is against vegan diets -> the author is against veganism} \newline
    stance: against \newline
    \newline
    tweet: <The rainbow flag means more than just a pride symbol. It's a symbol of our fight for EQUALITY. \#LoveIsLove> \newline
    target: LGBTQ Rights \newline
    {\color{blue}reasoning: the author is in favor of equal rights for the LGBTQ community -> the author is in favor of LGBTQ rights} \newline
    stance: favor \newline
    \newline
    tweet: <I love the way the sun sets every day. \#Nature \#Beauty> \newline
    target: Taxes \newline
    {\color{blue}reasoning: the author is in favor of nature and beauty -> the author is neutral towards taxes} \newline
    stance: none \newline
    \newline
    tweet: <@lifekingra @guardian The public can't be trusted to be 100\% honest in their "truthful" interpretations and memories.> \newline
    target: Police Body Camera Ban \newline
    {\color{blue}reasoning: the author is against relying on the public's interpretations and memories -> the author is against of police body camera ban} \newline
    stance: against \newline
    \newline
    tweet: <Veganism is not a restriction but rather an expansion of your love, care and respect for all creatures.> \newline
    target: Animal Rights \newline
    {\color{blue}reasoning: the author is in favor of veganism -> the author is in favor of animals -> the author is in favor of animal rights} \newline
    stance: favor \newline 
    \newline
    tweet: <There is no shortcut to success, there's only hard work and dedication \#Success \#SuccessMantra> \newline
    target: Open Borders \newline
    {\color{blue}reasoning: the author is in favor of hard work and dedication -> the author is neutral towards open borders} \newline
    stance: none \newline
\end{framed}

\end{document}